\documentclass{article}

\usepackage{PRIMEarxiv}

\usepackage[utf8]{inputenc} 
\usepackage[T1]{fontenc}    
\usepackage{hyperref}       
\usepackage{url}            
\usepackage{booktabs}       
\usepackage{amsfonts}       
\usepackage{nicefrac}       
\usepackage{microtype}      
\usepackage{lipsum}
\usepackage{float}
\usepackage{newunicodechar}
\newunicodechar{⁻}{\textsuperscript{-}}
\usepackage{fancyhdr}       
\usepackage{graphicx}       
\usepackage{comment}
\graphicspath{{media/}}     

\title{Towards Nepali-Language LLMs: Efficient GPT Training with a Nepali BPE Tokenizer}

\author{
  Adarsha Shrestha \\
  Khwopa College of Engineering \\
  Bhaktapur, Nepal \\
  \texttt{\href{mailto:kce078bct003@khwopa.edu.np}{kce078bct003@khwopa.edu.np}}
  \And
  Basanta Pokharel \\
  Khwopa College of Engineering \\
  Bhaktapur, Nepal \\
  \texttt{\href{mailto:kce078bct008@khwopa.edu.np}{kce078bct008@khwopa.edu.np}}
  \And
  Binit Shrestha \\
  Khwopa College of Engineering \\
  Bhaktapur, Nepal \\
  \texttt{\href{mailto:kce078bct009@khwopa.edu.np}{kce078bct009@khwopa.edu.np}}
  \And
  Smriti Adhikari \\
  Khwopa College of Engineering \\
  Bhaktapur, Nepal \\
  \texttt{\href{mailto:kce078bct048@khwopa.edu.np}{kce078bct048@khwopa.edu.np}}
  \And
  Dinesh Gothe \\
  Khwopa College of Engineering \\
  Bhaktapur, Nepal \\
  \texttt{\href{mailto:gothe.dinesh@khwopa.edu.np}{gothe.dinesh@khwopa.edu.np}}
}

\begin{document}
\maketitle
\begin{abstract}
Nepali, a low-resource language spoken by over 32 million people, continues to face challenges in natural language processing (NLP) due to its complex grammar, agglutinative morphology, and limited availability of high-quality corpora. Most efforts to date have centered on basic encoder architectures; they remain insufficient for Nepali-specific text generation. This study presents a GPT-2–based Nepali language model trained using several training strategies inspired by GPT-3, including optimized learning rate schedules, batch scaling, and architectural refinements. A custom 16k Byte-Pair Encoding (BPE) tokenizer was trained exclusively on Nepali text to ensure more consistent segmentation and improved input representation. The model was pretrained on a combined dataset comprising a 10.75GB cleaned NepBERTa corpus and additional web-scraped Nepali news articles. FlashAttention was integrated to reduce memory usage and stabilize training. After two epochs, the model achieved a training loss of 3.168177, a validation loss of 3.081982, and a final perplexity of 21.80, demonstrating its capability to generate coherent Nepali news-style text.
\end{abstract}

\keywords{Nepali Language Model, Low-Resource NLP, GPT-2, GPT-3, Byte-Pair Encoding (BPE) Tokenizer, Perplexity}

\section{Introduction}
Recent advancements in natural language processing (NLP) have significantly improved automated text understanding and generation, enabling models to process and generate coherent, contextually relevant language. However, most NLP research has focused on widely spoken languages such as English, leaving languages like Nepali largely underexplored. This underscores the pressing need for dedicated NLP solutions tailored specifically to the Nepali language.

Nepali is considered a low-resource language[\cite{rajan2021survey}, \cite{basu2020identification}], making NLP research for such languages a persistent challenge. Despite being spoken by more than thirty-two million people, Nepali remains significantly underrepresented in modern NLP research. Nepali features \textbf{complex grammar}, \textbf{agglutinative morphology}, flexible sentence structure, and diverse writing styles. These linguistic characteristics, coupled with the limited availability of large, clean text corpora, make the development of effective language models difficult. Although interest in Nepali NLP has grown in recent years, much of the existing work has concentrated on encoder-based architectures such as BERT\cite{devlin-etal-2019-bert}. As a result, comparatively little progress has been made toward building strong generative models capable of producing coherent long-form Nepali text.

The introduction of the \textbf{Transformer} architecture marked a major turning point in NLP research. Vaswani et al.\ \cite{vaswani2017attention} demonstrated that the self-attention mechanism can capture long-range dependencies while enabling efficient parallel computation. Subsequent developments such as GPT-2 \cite{radford2019language} and GPT-3 \cite{brown2020language} further showed the capabilities of decoder-only autoregressive models trained on large and diverse corpora. These models excel at next-token prediction and can perform various tasks without explicit task-specific fine-tuning. However, training them effectively requires significant computational resources and carefully designed optimization strategies.

Recent advances in training efficiency have helped make such models more accessible. One such development is \textbf{FlashAttention}\cite{dao2022flashattention}, which introduces an IO-aware attention mechanism that reduces memory usage and improves computational speed. Tokenizer design has also become increasingly important. While GPT-4 employs a vocabulary of approximately one hundred thousand tokens for multilingual capability \cite{achiam2023gpt}, building smaller vocabularies using BPE\cite{sennrich2016neural} tokenizer are often more practical and computationally less expensive for monolingual, resource-constrained settings like Nepali.

Research on Nepali language modeling has progressed steadily. Monolingual models such as \textbf{NepBERTa} \cite{timilsina2022nepberta} demonstrated that Nepali-specific pretraining can outperform multilingual alternatives. More recent works have explored GPT-2-based Nepali models\cite{thapa2024development}, showing promising results but also highlighting challenges in generating coherent long-form text. Building on these developments, this study introduces a GPT-2-based Nepali language model trained using several optimization strategies inspired by GPT-3. The model integrates FlashAttention for improved memory efficiency and employs a custom \textbf{BPE tokenizer} tailored to Nepali subword patterns. Training is performed on a cleaned corpus of 10.75 GB that combines the NepBERTa dataset with newly collected Nepali news articles, allowing the model to capture both traditional and contemporary language usage.

\section{Literature Review}
\label{sec:headings}
The rapid development of monolingual language models tailored for languages beyond English, driven by the growing need for linguistically and culturally specific NLP systems. Models such as FinBERT \cite{virtanen2019multilingual} for Finnish, GBERT \cite{chan2020german} for German, FlauBERT \cite{le2020flaubert} for French, Chinese BERT \cite{cui2021pre}, and NepBERTa \cite{timilsina2022nepberta} for Nepali  have demonstrated that customizing tokenizers, pretraining corpora, and architectural choices for a single language can significantly improve downstream performance. These advancements highlight the effectiveness of monolingual pretraining, particularly for languages with rich morphology or limited linguistic resources, where multilingual models often fall short.

Building on this progress, Transformer architecture introduced by Vaswani et al. marked a revolutionary shift in NLP by establishing the self-attention mechanism, which enabled improved modeling of long-range dependencies through parallelized training \cite{vaswani2017attention}. Radford et al. built upon this foundation with GPT-2, demonstrating that models trained on diverse text corpora could perform various NLP tasks without explicit fine-tuning \cite{radford2019language}. Brown et al. later presented GPT-3 with 175 billion parameters, showcasing remarkable few-shot learning capabilities, although significant computational challenges remained \cite{brown2020language}.

Flash Attention brought critical developments in computational efficiency by addressing quadratic memory scaling through IO-aware implementation \cite{dao2022flashattention}. This approach achieved up to 40\% memory reduction and doubled computation speed. While GPT-4's optimized BPE tokenizer with approximately 100,000 tokens improved multilingual performance \cite{achiam2023gpt}, smaller vocabulary sizes offer distinct advantages for monolingual and resource-limited settings. These smaller tokenizers reduce model size and computational requirements while maintaining adequate linguistic coverage.

Nepali language processing has seen significant progress. NepBERTa \cite{timilsina2022nepberta} demonstrated that monolingual models trained on 0.8 billion words could outperform multilingual alternatives. Thapa et al. recently presented work on pretrained transformers using a 27.5 GB Nepali corpus, showing promise with instruction-tuned GPT-2 models \cite{thapa2024development}. However, challenges with long-form generation and computational costs remained. The integration of GPT-2 architecture with efficient BPE tokenization and Flash Attention's memory-efficient computation provides a practical foundation for advancing Nepali text generation in resource-constrained environments.

\section{Methodology}

\subsection{Dataset}

\subsubsection{Data Collection}
The dataset for this study combines the NepBERTa corpus with additional web-scraped Nepali news articles. NepBERTa, which was collected from 36 major Nepali news websites, provides a large and diverse source of written Nepali text \cite{timilsina2022nepberta}. However, because much of the dataset is several years old, it was supplemented with fresh articles from popular news portals such as Ekantipur, Setopati, Ratopati, and OnlineKhabar. These newer sources helped introduce recent vocabulary and writing styles that are useful for training modern language models. The OSCAR Nepali dataset \cite{Ortiz_Su_rez_2020} was examined but ultimately excluded because it contained a high amount of repeated and overlapping content. 

\subsubsection{Data Pre-Processing}
Before training, the combined dataset was cleaned to remove noise and ensure consistent text quality. Regex-based filtering was applied to strip out HTML tags, JavaScript code, URLs, English words, and any characters not written in Devanagari script. All text was normalized, and duplicate lines, repeated headlines, incomplete fragments, and unusually long or short sentences were removed. The cleaned NepBERTa text and the crawled news articles were then merged into a single file to create a unified training corpus. After merging and cleaning all usable text, the final corpus amounted to 10.75 GB.

After cleaning, a custom Byte-Pair Encoding (BPE) tokenizer \cite{sennrich2016neural} was trained with the vocabulary size of 16,384 using the SentencePiece \cite{kudo2018sentencepiecesimplelanguageindependent} library. The tokenizer was trained on an input sample of 8 million characters, with a maximum sentence length of 8192 characters and a character coverage of 0.9995. This tokenizer was then used to convert the full dataset into tokenized shards stored as NumPy files. The entire corpus was structured into 87 shards, each containing approximately 10 million tokens, enabling efficient loading during model training.

\begin{figure}[h]
	\centering
	\includegraphics[width=0.4\textwidth]{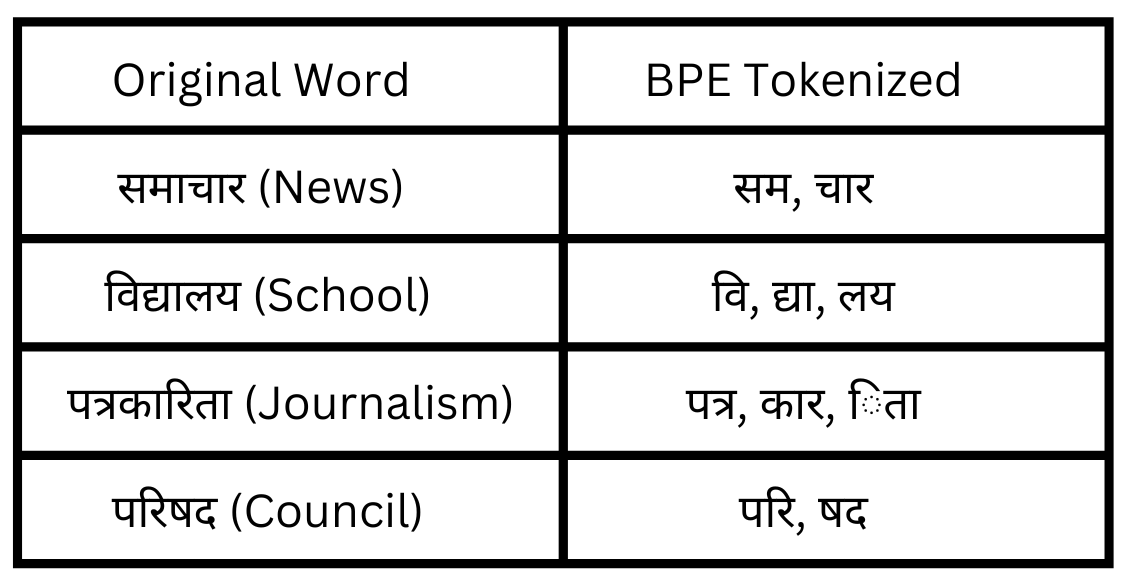}
	\caption{Example of BPE tokenization in Nepali
    }
\end{figure}

\subsection{Model Architecture and Hyperparameters}

Our model follows the decoder-only transformer architecture introduced in GPT-2 \cite{radford2019language}. It preserves the same structural configuration: 12 decoder layers, 12 attention heads, and an embedding size of 768. Unlike the original GPT-2, which used a vocabulary size of 50,257, our model employs a vocabulary size of 16,384 optimized for Nepali subword patterns. This adjustment reduces the total parameter count to approximately 98 million.Training a transformer with a large effective batch size is difficult under limited GPU memory. To address this, micro-batches of size 8 was applied and gradient were accumulated for 64 steps, resulting in an effective batch size of 524,288 tokens. The model is trained with a sequence length of 1024, using the same micro-batch size for both training and evaluation. A total of 2 epochs were run, corresponding to 3300 training steps. To further improve training efficiency, Flash Attention \cite{dao2022flashattention} was incorporated, which computes attention in tiled blocks rather than storing large intermediate matrices. This significantly reduces memory usage and increases throughput, enabling stable training even with large effective batch sizes. Most hyperparameters follow commonly used GPT-3 \cite{brown2020language} training configurations. The weight decay is set to 0.1 to prevent overfitting. Weight decay is a regularization technique that helps to prevent overfitting by penalizing large weights. It
is particularly effective in models like neural networks, where large weights can
lead to poor generalization \cite{loshchilov2017decoupled}. The maximum learning rate is set to $6\times10^{-4}$, with a minimum learning rate of $6\times10^{-5}$. To improve memory efficiency while preserving model quality, training uses \texttt{bfloat16} precision. AdamW optimizer was used with $\beta_{1}=0.9$, $\beta_{2}=0.95$, and $\epsilon=1\mathrm{e}{-8}$. A learning-rate warm-up for the first 715 steps was applied, gradually increasing the learning rate from near zero to the peak value. After warm-up, a cosine decay schedule reduces the learning rate smoothly for the rest of training \cite{smith2017cyclical}.

\section{Evaluation}

To assess the performance of the model, cross-entropy loss and perplexity were used as primary evaluation metrics. Cross-entropy reflects how accurately the model predicts the next token, whereas perplexity measures how confidently the model generates text based on preceding context\cite{morgan2024perplexity}. Lower values for both metrics indicate better language modeling performance.
During training, the model achieved a training loss of 3.168 and a validation loss of 3.081, demonstrating stable learning without signs of overfitting. The resulting perplexity score of 21.80 indicates that the model can effectively predict Nepali text sequences with reasonable confidence. Given the linguistic complexity and low-resource nature of the Nepali language, this perplexity reflects strong performance for a GPT-2–based architecture trained on a limited but diverse corpus.

Overall, the evaluation results confirm that the model generalizes well and is capable of generating coherent and contextually relevant Nepali text across multiple domains.

\section{Results}
The model was trained for a total of 2 epochs, corresponding to 3300 training steps, using the custom tokenizer. Table 1 shows selected training and validation metrics, while Figures 1 and 2 illustrate the progression of cross-entropy loss and perplexity. Both loss and perplexity steadily decrease over time, indicating consistent convergence, with the final validation perplexity reaching ~21.8, demonstrating the model’s capability to generate coherent Nepali text.

\begin{table}[htbp]
\centering
\caption{Key Training and Validation Metrics}
\begin{tabular}{|c|c|c|c|}
\hline
\textbf{Steps} & \textbf{Train Loss} & \textbf{Val Loss} & \textbf{Perplexity} \\
\hline
0    & 9.8422 & 9.8449 & 18862.37 \\
500  & 5.5788 & 5.4479 & 232.28   \\
1000 & 4.6346 & 4.4944 & 89.52    \\
1500 & 3.2993 & 3.7757 & 43.63    \\
2000 & 3.6195 & 3.4703 & 32.15    \\
2500 & 3.5218 & 3.2102 & 24.79    \\
3000 & 3.5967 & 3.1450 & 23.22    \\
3299 & 3.1682 & 3.0820 & 21.80    \\
\hline
\end{tabular}
\label{tab:training_metrics}
\end{table}

\begin{figure}[H]
	\centering
	\includegraphics[width=0.55\textwidth]{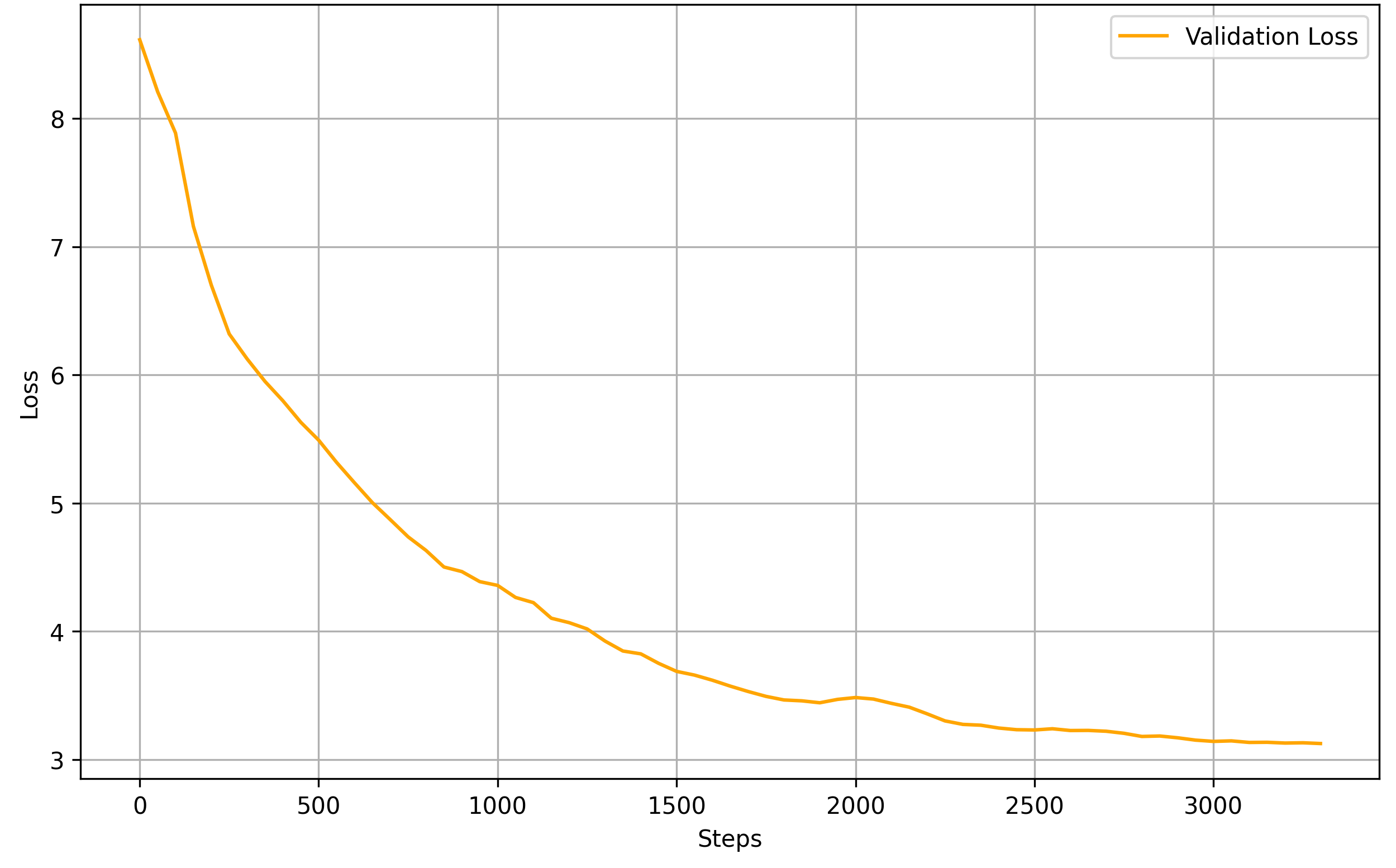}
	\caption{Cross-Entropy Loss vs. Steps for Training and Validation
    }
\end{figure}

\begin{figure}[H]
	\centering
	\includegraphics[width=0.55\textwidth]{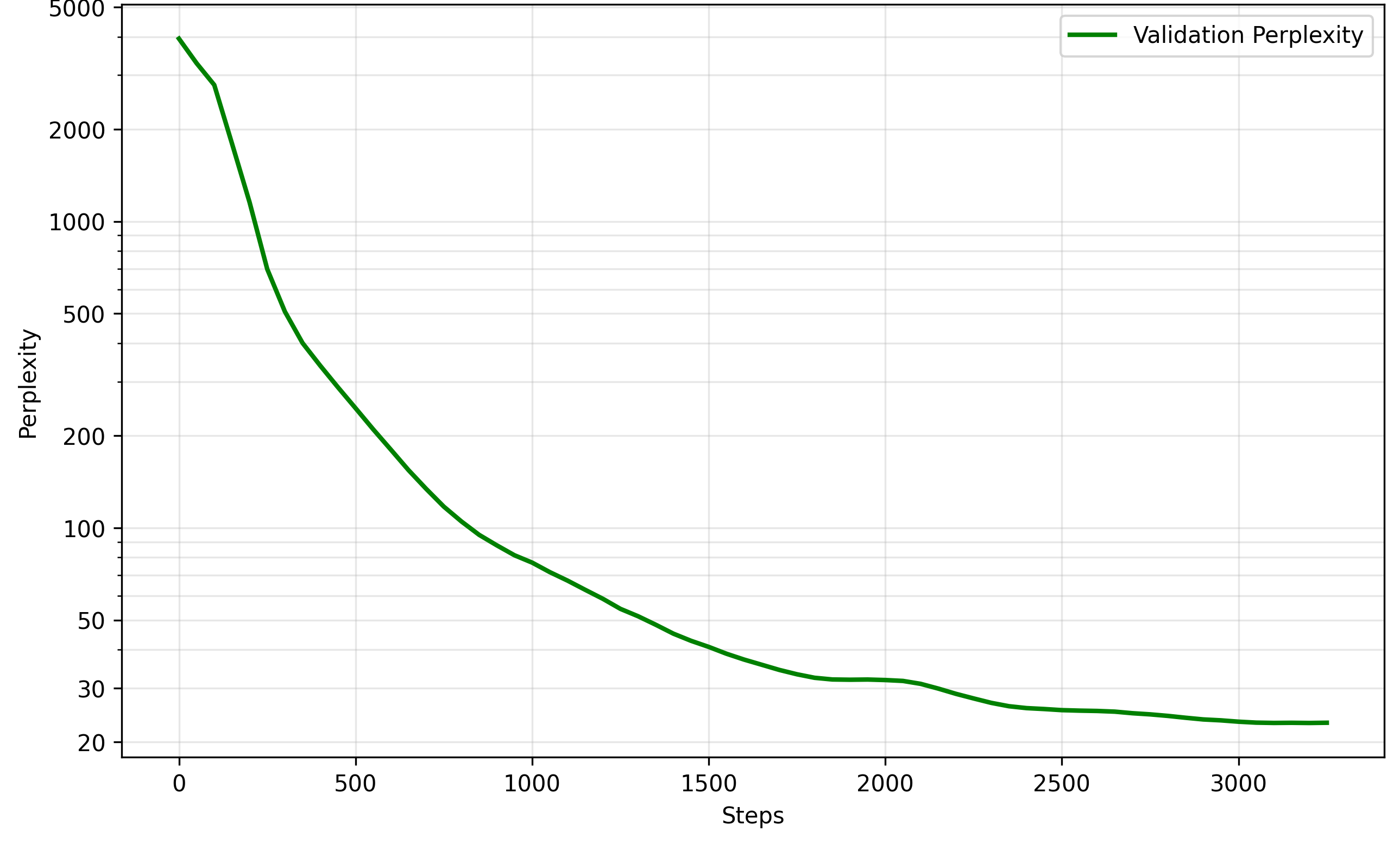}
	\caption{Perplexity vs. Steps for Training and Validation
    }
\end{figure}

\section{Conclusion}
In conclusion, we successfully developed a Nepali language model achieving a perplexity of 21.80. The model is capable of generating text that is accurate, consistent, and coherent based on human evaluation. The Nepali Corpus and NepaliGPT datasets created during this research can serve as valuable resources for improving future generative language models. Further enhancements can be achieved by using larger and more diverse datasets, which would enable the model to understand varied terminologies and contexts. Additionally, fine-tuning the model on instruction-based datasets may improve its ability to follow specific instructions and generate task-oriented Nepali text more effectively.

\section*{Acknowledgments}
We express our sincere gratitude to our supervisors, Er. Nirajan Bekoju and Er. Dinesh Gothe, for their expert guidance, invaluable feedback, and continuous support throughout this project. Their insights into natural language processing greatly strengthened the quality and direction of our work.

We also extend our appreciation to our college for providing the necessary resources, including GPU-enabled computational facilities, which were essential for training and refining our model.

\bibliographystyle{unsrt}  
\bibliography{references}

\end{document}